\documentclass{svproc}
\usepackage{url}

\usepackage{hyperref}
\usepackage{algorithm}
\usepackage{algorithmic}
\usepackage{textcomp}
\usepackage{float}
\usepackage{algorithm}
\usepackage{algorithmic}
\usepackage{mathtools}
\usepackage{commath}
\usepackage{fixmath}
\usepackage{subcaption}
\usepackage{import}
\usepackage{bm,bbm}
\usepackage[strings]{underscore}
\usepackage{setspace}
\let\Algorithm\algorithm
\renewcommand\algorithm[1][]{\Algorithm[#1]\setstretch{1.15}}
\usepackage{textcomp}

\usepackage[mathscr]{euscript}
\usepackage[bbgreekl]{mathbbol}
\usepackage{stfloats}

\usepackage{enumitem}
\setlist[itemize,1]{label=\textbullet}

\DeclareMathAlphabet{\mathbbold}{U}{bbold}{m}{n}

\def \hrsup_size{0.48}
\def \framesu_size{0.48}

\begin{document}
\mainmatter              
\title{Towards Partner-Aware Humanoid Robot Control under Physical Interactions}
%
%

\author{Yeshasvi Tirupachuri\inst{1}\inst{2}\inst{3} \and Gabriele Nava\inst{2}\inst{3}
\and Claudia Latella\inst{2} \and Diego Ferigo\inst{2}\inst{4}
\and \\ Lorenzo Rapetti\inst{2}\inst{4} \and Luca Tagliapietra\inst{2}
\and Francesco Nori\inst{5} \and Daniele Pucci\inst{2}}

%
%
%

\institute{RBCS, Istituto Italiano di Tecnologia, Genova, Italy
           \and
           Dynamic Interaction Control, Istituto Italiano di Tecnologia, Genova, Italy
           \and
           DIBRIS, University of Genova, Genova, Italy
           \and
           School of Computer Science, University of Manchester, Manchester, United Kingdom
           \and
           DeepMind, London, United Kingdom}

\maketitle              

\begin{abstract}
The topic of physical human-robot interaction received a lot of attention from the robotics community because of many promising application domains. However, studying physical interaction between a robot and an external agent, like a human or another robot, without considering the dynamics of both the systems may lead to many shortcomings in fully exploiting the interaction. In this paper, we present a coupled-dynamics formalism followed by a sound approach in exploiting helpful interaction with a humanoid robot. In particular, we propose the first attempt to define and exploit the human help for the robot to accomplish a specific task. As a result, we present a task-based partner-aware robot control techniques. The theoretical results are validated by conducting experiments with two iCub humanoid robots involved in physical interaction.


\end{abstract}

\section{INTRODUCTION}
\label{introduction}

The evolution of robotic systems over the last decade is much more rapid than it has ever been since their debut. Robots existed as separate entities till now but the horizons of a symbiotic human-robot partnership are impending. In particular, application domains like elderly care, collaborative manufacturing, collaborative manipulation, etc., are considered the need of the hour. Across all these domains, it is crucial for robots to physically interact with humans to either assist them or to augment their capabilities. Such \emph{human in the loop} physical human-robot interaction (pHRI) scenarios demand careful consideration of both the human and the robot systems while designing controllers to facilitate robust interaction strategies for successful task completion. More importantly, a generalized human-robot interaction formalism is needed to study the physical interaction adaptation and exploitation. Towards that goal, in this paper, we present a generalized human-robot interaction formalism and partner-aware robot control techniques.

The three main components of any pHRI scenario are: 1) a \textit{robotic agent}, 2) a \textit{human agent} and, 3) the \textit{environment} surrounding them. Over the course of time, physical interactions are present between any of the two components. An intuitive conceptual representation of the interactions occurring during pHRI is presented in \cite{losey2018review}. More specifically, the interaction between a human and a humanoid robot is particularly challenging because of the complexity of the robotic system \cite{goodrich2008human}. Unlike traditional industrial robots which are fixed base by design, humanoid robots are designed as floating base systems to facilitate anthropomorphic navigational capabilities. The aspect of balancing has received a lot of attention in the humanoid robotics community and several prior efforts  \cite{caux1998balance} \cite{hirai1998development} \cite{hyon2007full} went into building controllers that ensure stable robot behavior. More recently momentum-based control proved to be a robust approach and several successful applications have been realized \cite{stephens2010dynamic} \cite{herzog2014balancing} \cite{koolen2016design} \cite{hofmann2009exploiting} ensuring contact stability \cite{nori2015icub} with the environment by monitoring contact wrenches through quadratic programming \cite{ott2011posture} \cite{wensing2013generation} \cite{nava2016stability}. In general, these controllers are built to ensure robustness to any external perturbations and hence they are often blind to any helpful interaction a human is trying to have with the robot to help achieve its task.

The knowledge of human intent is a key element for successful realization of pHRI tasks. The process of human intent detection is broadly divided into \textit{intent information}, \textit{intent measurement} and \textit{intent interpretation} \cite{losey2018review}. The choice of a communication channel is directly related to intent measurement and affects the robot's ability to understand human intent. Accordingly, a myriad of technologies have been used as interfaces for different applications of pHRI. In the context of rehabilitation robotics, electroencephalography (EEG) \cite{mattar2018biomimetic} \cite{sarac2013brain} \cite{mcmullen2014demonstration} and electromyography (EMG) \cite{radmand2014characterization} \cite{au2008powered} \cite{song2008assistive} \cite{zhou2018multi} proved to be invaluable. Force myography (FMG) \cite{cho2016force} \cite{yap2016design} \cite{rasouli2016towards} is a relatively new technology which has been successfully used in rehabilitation. EMG has also been successfully used by \cite{peternel2018robot} to realize a pHRI collaborative application to continuously monitor human fatigue. Force/Torque sensors mounted on the robots are often the most relied technology in pHRI scenarios for robot control as they facilitate direct monitoring and regulation of the interaction wrenches between the human and the robot \cite{bussy2012human} \cite{bussy2012proactive} \cite{ikemoto2012physical} \cite{peternel2013learning} \cite{donner2016cooperative}. Vision based techniques like human skeletal tracking \cite{reily2018skeleton}, human motion estimation \cite{kyrkjebo2018inertial} and hand gesture recognition \cite{rautaray2015vision} are also used as interfaces.

In general, the designer decides on the choice of the interface, to communicate the human intent, depending on the application and often times using a single interface mode is limiting. Hence, a combination of vision and haptic interfaces are used in literature to realize successful applications of human-robot collaboration \cite{agravante2014collaborative} \cite{de2007human}. However, we believe there is an impending change in this paradigm and the future technologies of pHRI will leverage on getting as much holistic information as possible from humans involved in pHRI, especially for domains like collaborative manufacturing. Having both the kinematic quantities like joint positions and velocities and dynamic quantities like joint accelerations and torques of the human will enable real-time monitoring of the human dynamics to build robust controllers for successful task completion taking into account the physical interactions between the human and the robot.

In a typical physical interaction scenario, the dynamics of the two agents involved play a crucial role in shaping the interaction. So, in order to understand the interaction more concretely, the dynamics of both the systems have to be considered together rather than in isolation. In this paper, we take into account the dynamics of the combined system and present a \emph{coupled-dynamics formalism}. Also, we attempt at mathematically characterizing and quantifying external helpful interaction with the robot that contributes towards task completion. Furthermore, we present a task-based partner-aware robot control techniques that account for external help. We validate our approach using an experimental scenario in which assistance provided by an external interacting agent is leveraged by the robot for its task completion. This paper is organized as follows: Section \ref{background} introduces the notations, the system modeling, and the typical contact constraints. Section \ref{control} presents the task-based control law. Section \ref{experiments} lay the details of the experiments conducted. Section \ref{results} presents the results followed by conclusions and further extensions.

\section{BACKGROUND}
\label{background}

\subsection{Notation}
In this section, we present the basic notation used in this paper. \textit{A} denotes an inertial frame, with \textit{z}-axis pointing against the gravity. The constant \textit{g} denotes the norm of the gravitational acceleration. We advise the reader to pay close attention to the notational nuances. The human-related notations are denoted with $\mathbbold{double-bold}$ terms, e.g., $\mathbbold{n}$, robot-related notations are denoted with ``$\mathrm{straight}$'' terms, e.g., $\mathrm{n}$, and notations that apply to both the systems are denoted with $slanted$ terms, e.g., $n$. In addition, composite matrices are denoted with $\mathbold{BOLD}$ terms, e.g., $\mathbold{M}$.

\subsection{Modeling}

\begin{figure}[ht]
	\centering
	\includegraphics[scale=0.35]{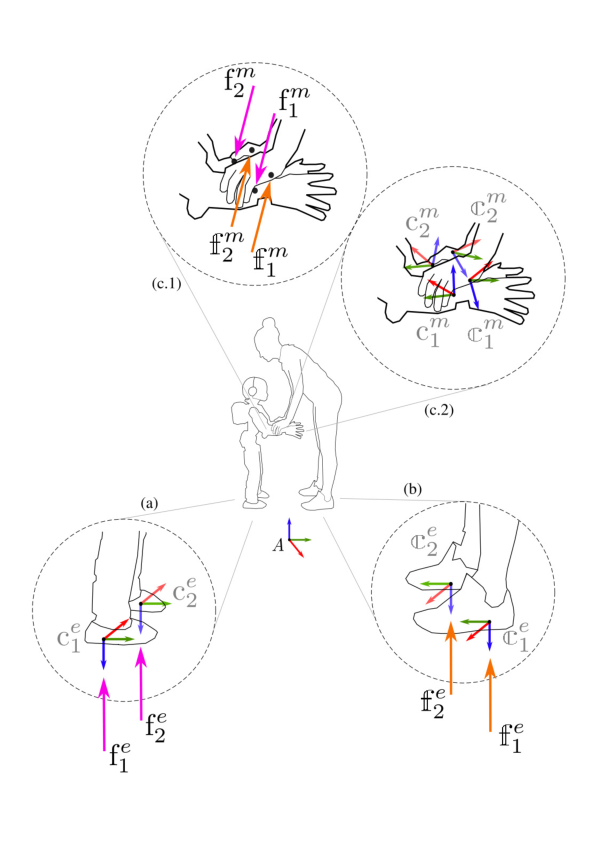}
	\caption{A typical human-robot dynamic interaction scenario}
	\label{pHRI}
\end{figure}

A typical physical human-robot interaction scenario is shown in Fig. \ref{pHRI}. There are two agents: the human, and the robot. Both agents are physically interacting with the environment and, in addition, are also engaged in physical interaction with each other.

In the first approximation, both the human and the robot can be considered as multi-body mechanical systems composed of  $\mathbbold{n}+1$ and $\mathrm{n}+1$ rigid bodies respectively, called links, connected through $\mathbbold{n} \in \mathbb{N}$ and $\mathrm{n} \in \mathbb{N}$ joints with one degree of freedom. Even though the assumption of a human body being modeled as rigid bodies is far from reality, it serves as a rough approximation when formulating physical human-robot interaction dynamics and allows us to synthesize robot controllers optimizing both human and robot variables. Further, we consider both the human and the robot to be \textit{free-floating} systems, i.e. none of the links have an \textit{a priori} constant pose with respect to the inertial frame.

The configuration space of a \textit{free-floating} system is characterized by the \textit{joint positions} and the pose of a specific \textit{frame} attached to a link of the system, generally referred to as \textit{base frame} denoted by $B$. In the case of a \textit{free-floating} mechanical system, the configuration space is a set of elements representing the \textit{floating base} and the total number of joints, say $n$. Hence, it lies on the Lie group $\mathbb{Q} = \mathbb{R}^3 \times SO(3) \times \mathbb{R}^n$. We denote an element in the configuration space with $q = (q_b,s) \in \mathbb{Q}$. It consists of pose of the \textit{base frame} $q_b = (^{\scalebox{\framesu_size}{A}}p_{\scalebox{\framesu_size}{B}}, ^{\scalebox{\framesu_size}{A}}R_{\scalebox{\framesu_size}{B}}) \in \mathbb{R}^3 \times SO(3)$ where  $^{\scalebox{\framesu_size}{A}}p_{\scalebox{\framesu_size}{B}} \in \mathbb{R}^3$ denotes the position of the base frame with respect to the inertial frame; $^{\scalebox{\framesu_size}{A}}R_{\scalebox{\framesu_size}{B}} \in SO(3)$ denotes the rotation matrix representing the orientation of the base frame with respect to the inertial frame; and the joint positions vector $s \in \mathbb{R}^n$ capturing the topology (\textit{shape}) of the robot. Specifically, $\mathbbold{q} \in \mathbb{Q}$ denotes an element of the human configuration space and $\mathrm{q} \in \mathbb{Q}$ denotes an element of the robot configuration space.

To the purpose of finding mathematical models for both the (approximation of the) human and the robot, we apply the Euler-Poincar\'{e} formalism to both multi-body systems~\cite{Marsden2010}. Then, we obtain the following equations of motion describing the dynamics of the human and the robotic agents respectively:

\begin{subequations}
\begin{equation}
	\mathbbold{M}(\mathbbold{q}) \dot{\bbnu} + \mathbbold{C}(\mathbbold{q},\bbnu)\bbnu + \mathbbold{G}(\mathbbold{q}) =
	 \mathbbold{B} {\tau}^{\scalebox{\hrsup_size}{H}} + \mathbbold{J}^T\mathbbold{f}^*
	\label{NEHuman2}
\end{equation}
\begin{equation}
	\mathrm{M}(\mathrm{q}) \dot{\nu} + \mathrm{C}(\mathrm{q},\nu)\nu + \mathrm{G}(\mathrm{q}) =
	\mathrm{B} {\tau}^{\scalebox{\hrsup_size}{R}} + \mathrm{J}^T \mathrm{f}^*
	\label{NERobot2}
\end{equation}
\label{NE}
\end{subequations}

In general, $M \in \mathbbold{R}^{n+6 \times n+6}$ is the mass matrix, $C \in \mathbb{R}^{n+6 \times n+6}$ is the Coriolis matrix, $G \in \mathbb{R}^{n+6}$ is the gravity term, $B = (0_{n \times 6},1_n)^T$ is a selector matrix, ${\tau}^{\scalebox{\hrsup_size}{H,R}}  \in \mathbb{R}^{n}$ is a vector representing the agent joint torques. Here, $n$ is the number of joints that are assumed to compose either the model of the human or that of the robot, and may be different in the two cases. We also assume that each agent, is subject to a total of $n_c = n_m+n_e \in \mathbb{N}$ distinct wrenches. These wrenches are composed of two subsets: the wrenches due to \textit{mutual} interaction, denoted with the subscript $m$ and the wrenches exchanged between the \textit{agent} and the \textit{environment}, denoted with the subscript $e$ respectively, In either case, the contact wrenches are represented by:

\begin{equation}
	f^* = \begin{bmatrix}
		 	f^{\scalebox{\hrsup_size}{\textit{m}}}_1; &&
		 	f^{\scalebox{\hrsup_size}{\textit{m}}}_2; && .... && f^{\scalebox{\hrsup_size}{\textit{m}}}_{n_m}; && \\ \\
		 	f^{\scalebox{\hrsup_size}{\textit{e}}}_{{n_e}+1}; &&
		 	f^{\scalebox{\hrsup_size}{\textit{e}}}_{{n_e}+2}; && .... && f^{\scalebox{\hrsup_size}{\textit{e}}}_{n_e}
	      \end{bmatrix} \in \mathbb{R}^{6{n_c}} \notag
\end{equation}

Accordingly,
$\mathbbold{f}^* = \begin{bmatrix}
			        \mathbbold{f}_m && \mathbbold{f}_e
                 \end{bmatrix}^T$
				 with $\mathbbold{f}_m$ the external wrenches applied on the human agent by the robotic agent and
				      $\mathbbold{f}_e$ the external wrenches applied on the human agent by the environment. Similarly, $\mathrm{f}^*= \begin{bmatrix}
							\mathrm{f}_m && \mathrm{f}_e
						  \end{bmatrix}^T$
						with $\mathrm{f}_m$ the external wrenches applied on the robotic agent by the human agent and
						     $\mathrm{f}_e$ the external wrenches applied on the robotic agent by the environment.

We define a set of frames
$\mathscr{C} = \{c_1,c_2,....c_{n_m},c_{n_{m+1}},c_{n_{m+2}},....,c_{n_e} \}$
and assume that the application point of the $k$-th external wrench on an agent is associated with a frame $c_k \in \mathscr{C}$, attached to the agent link on which the wrench acts, and has \textit{z}-axis pointing in the direction normal to the contact plane. Furthermore, the external wrench $f^{\scalebox{\hrsup_size}{m/e}}_k$ is expressed in a frame whose orientation is that of the inertial frame $A$, and whose origin is that of the frame $c_k$.

The jacobian $J_{c_k} = J_{c_k}(q)$ is the map between the agent's velocity
$\nu = [\nu_B; \dot{s}]\in \mathbb{R}^{n+6}$
and the velocity of the frame $c_k$ given by $^{A}v_{c_k} = [ ^{A}\dot{p}_{c_k}; \ ^{A}{{\omega}_{c_k}} ]$:

\begin{equation}
	^{A}{v}_{c_k} = J_{c_k}{\nu}
	\label{EEVelocity}
\end{equation}
The jacobian matrix has the following structure~\cite{Featherstone2007}:

\begin{subequations}
\begin{equation}
	{J_{c_k}}(q) = \begin{bmatrix}
						{J^b_{c_k}}(q) && {J^j_{c_k}}(q)
				   \end{bmatrix} \in \mathbb{R}^{6 \times n+6}
\end{equation}

\begin{equation}
	{J^b_{c_k}}(q) = \begin{bmatrix}
						1_3 && -S({^{A}p_{c_k}} - {^{A}p_B}) \\
						0_{3 \times 3} && 1_3
					\end{bmatrix} \in \mathbb{R}^{6 \times 6}
\end{equation}
\end{subequations}

\subsection{Contact Constraints}

We assume that holonomic constraints of the form $c(q) = 0$ act on both the human and the robot during their interaction with the environment. The links that are in contact with the ground can be considered as end-effector links that are rigidly fixed to the ground for the duration of the contact and hence have zero velocity. Following the equation (\ref{EEVelocity}), this can be represented as follows for the human and the robot respectively:

\begin{subequations}
	\begin{equation}
		\mathbbold{J}_{\mathbbold{c}_k}	\bbnu = 0
	\end{equation}
	\begin{equation}
		\mathrm{J}_{\mathrm{c}_k} \nu = 0
	\end{equation}
\end{subequations}
Differentiating the above kinematic constraints yields:
\begin{subequations}
	\begin{equation}
		\begin{bmatrix}
			\mathbbold{J}^b_{\mathbbold{c}_k} && \mathbbold{J}^j_{\mathbbold{c}_k}
		\end{bmatrix}
		\begin{bmatrix}
			\dot{\mathbbold{v}}_B \\
			\ddot{\mathbbold{s}}
		\end{bmatrix} +
		\begin{bmatrix}
			\dot{\mathbbold{J}}^b_{\mathbbold{c}_k} && \dot{\mathbbold{J}}^j_{\mathbbold{c}_k}
		\end{bmatrix}
		\begin{bmatrix}
			\mathbbold{v}_B \\
			\dot{\mathbbold{s}}
		\end{bmatrix} = 0
		\label{holcon1}
	\end{equation}
	\begin{equation}
		\begin{bmatrix}
			\mathrm{J}^b_{\mathrm{c}_k} && \mathrm{J}^j_{\mathrm{c}_k}
		\end{bmatrix}
		\begin{bmatrix}
			\dot{\mathrm{v}}_B \\
			\ddot{\mathrm{s}}
		\end{bmatrix} +
		\begin{bmatrix}
			\dot{\mathrm{J}}^b_{\mathrm{c}_k} && \dot{\mathrm{J}}^j_{\mathrm{c}_k}
		\end{bmatrix}
		\begin{bmatrix}
			\mathrm{v}_B \\
			\dot{\mathrm{s}}
		\end{bmatrix} = 0
		\label{holcon2}
	\end{equation}
	\label{holcon}
\end{subequations}

Now, during physical human-robot interaction, there is a contact between the robot and the human. We assume that these contacts can be modeled as holonomic constraints of the form $c(\mathbbold{q},\mathrm{q}) = 0$. To this purpose, we consider a frame $\mathbbold{c}_m^{\scalebox{\hrsup_size}{H}} \in \mathscr{C}^{\scalebox{\hrsup_size}{H}}$ attached to the human link which is in contact with the robot. More precisely, let $^{A}T_{\mathbbold{c}_m^{\scalebox{\hrsup_size}{H}}}(\mathbbold{q})$ denote the homogeneous transformation from $\mathbbold{c}_m^{\scalebox{\hrsup_size}{H}}$ to the inertial frame. Similarly, we consider another frame $\mathrm{c}_m^{\scalebox{\hrsup_size}{R}} \in \mathscr{C}^{\scalebox{\hrsup_size}{R}}$ attached to the robot link in contact with the human. Let $^{A}T_{\mathrm{c}_m^{\scalebox{\hrsup_size}{R}}}(\mathrm{q})$ denote the homogeneous transformation from $\mathrm{c}_m^{\scalebox{\hrsup_size}{R}}$ to the inertial frame. The relative transformation between the frames $\mathbbold{c}_m^{\scalebox{\hrsup_size}{H}}$ and $\mathrm{c}_m^{\scalebox{\hrsup_size}{R}}$ is given by:

\begin{equation}
	^{\mathbbold{c}_m^{\scalebox{\hrsup_size}{H}}}T_{\mathrm{c}_m^{\scalebox{\hrsup_size}{R}}} =
	{^{A}T^{-1}_{\mathbbold{c}_m^{\scalebox{\hrsup_size}{H}}}}(\mathbbold{q}) \
	^{A}T_{\mathrm{c}_m^{\scalebox{\hrsup_size}{R}}}({\mathrm{q}})
	\label{HRcon}
\end{equation}

When $^{\mathbbold{c}_m^{\scalebox{\hrsup_size}{H}}}T_{\mathrm{c}_m^{\scalebox{\hrsup_size}{R}}}$ (or a part of it) is constant, it means that the robot and the human are in contact. By setting $^{\mathbbold{c}_m^{\scalebox{\hrsup_size}{H}}}T_{\mathrm{c}_m^{\scalebox{\hrsup_size}{R}}}$  to a constant, we obtain the aforementioned holonomic constraint of the form $c(\mathbbold{q},\mathrm{q}) = 0$. We assume a stable contact between the human and the robot during physical human-robot interaction, which leads to the condition that the relative velocity between the two frames $\mathbbold{c}_m^{\scalebox{\hrsup_size}{H}}$ and $\mathrm{c}_m^{\scalebox{\hrsup_size}{R}}$ is zero, i.e., the two contact frames move with the same velocity with respect to the inertial frame as given by the following relation:

\begin{equation}
	^{A}{\mathbbold{v}}_{\mathbbold{c}_m^{\scalebox{\hrsup_size}{H}}} = \
	{^{{\mathbbold{c}_m^{\scalebox{\hrsup_size}{H}}}}{X}_{\mathbbold{c}_m^{\scalebox{\hrsup_size}{R}}}} \ ^{A}{\mathrm{v}}_{\mathrm{c}_m^{\scalebox{\hrsup_size}{R}}}
	\label{HRholcongeneral}
\end{equation}
where ${^{{\mathbbold{c}_m^{\scalebox{\hrsup_size}{H}}}}{X}_{\mathbbold{c}_m^{\scalebox{\hrsup_size}{R}}}}$ is a frame transformation matrix. In this work we assume the contact frames to be coinciding and hence, the transformation matrix is Identity i.e. ${^{{\mathbbold{c}_m^{\scalebox{\hrsup_size}{H}}}}{X}_{\mathbbold{c}_m^{\scalebox{\hrsup_size}{R}}}} = I_{6 \times 6}$

\vspace{0.3cm}
In light of the above, the equation (\ref{HRholcongeneral}) can be written as:

\begin{equation}
	^{A}{\mathbbold{v}}_{\mathbbold{c}_m^{\scalebox{\hrsup_size}{H}}} = \
	^{A}{\mathrm{v}}_{\mathrm{c}_m^{\scalebox{\hrsup_size}{R}}},
	\label{HRholcon-specific}
\end{equation}

which can be represented as follows
\begin{equation}
	\mathbbold{J_{\mathbbold{c}_m^{\scalebox{\hrsup_size}{H}}}} \bbnu =
	\mathrm{J_{\mathrm{c}_m^{\scalebox{\hrsup_size}{R}}}} \nu
	\label{HRholcon}
\end{equation}

Differentiating the equation (\ref{HRholcon}) we get,
\begin{subequations}
\begin{equation}
	\begin{bmatrix}
		\mathbbold{J}^b_{\mathbbold{c}_m^{\scalebox{\hrsup_size}{H}}} && \mathbbold{J}^j_{\mathbbold{c}_m^{\scalebox{\hrsup_size}{H}}}
	\end{bmatrix}
	\begin{bmatrix}
		\dot{\mathbbold{v}}_B \\
        \ddot{\mathbbold{s}}
	\end{bmatrix} +
	\begin{bmatrix}
		\dot{\mathbbold{J}}^b_{\mathbbold{c}_m^{\scalebox{\hrsup_size}{H}}} && \dot{\mathbbold{J}}^j_{\mathbbold{c}_m^{\scalebox{\hrsup_size}{H}}}
	\end{bmatrix}
	\begin{bmatrix}
		\mathbbold{v}_B \\
        \dot{\mathbbold{s}}
	\end{bmatrix}  = \notag
\end{equation}

\begin{equation}
	\begin{bmatrix}
		\mathrm{J}^b_{\mathrm{c}_m^{\scalebox{\hrsup_size}{R}}} && \mathrm{J}^j_{\mathrm{c}_m^{\scalebox{\hrsup_size}{R}}}
	\end{bmatrix}
	\begin{bmatrix}
		\dot{\mathrm{v}}_B \\
        \ddot{\mathrm{s}}
	\end{bmatrix} +
	\begin{bmatrix}
		\dot{\mathrm{J}}^b_{\mathrm{c}_m^{\scalebox{\hrsup_size}{R}}} && \dot{\mathrm{J}}^j_{\mathrm{c}_m^{\scalebox{\hrsup_size}{R}}}
	\end{bmatrix}
	\begin{bmatrix}
		\mathrm{v}_B \\
        \dot{\mathrm{s}}
	\end{bmatrix}
	\label{HRholcon1}
\end{equation}

\begin{equation}
	\begin{bmatrix}
		\dot{\mathbbold{J}}^b_{\mathbbold{c}_m^{\scalebox{\hrsup_size}{H}}} && \dot{\mathbbold{J}}^j_{\mathbbold{c}_m^{\scalebox{\hrsup_size}{H}}} &&
		-\dot{\mathrm{J}}^b_{\mathrm{c}_m^{\scalebox{\hrsup_size}{R}}} && -\dot{\mathrm{J}}^j_{\mathrm{c}_m^{\scalebox{\hrsup_size}{R}}}
	\end{bmatrix}
	\begin{bmatrix}
		\mathbbold{v}_B \\
        \dot{\mathbbold{s}} \\
        \mathrm{v}_B \\
        \dot{\mathrm{s}}
	\end{bmatrix} + \notag
\end{equation}

\begin{equation}
	\begin{bmatrix}
		\mathbbold{J}^b_{\mathbbold{c}_m^{\scalebox{\hrsup_size}{H}}} && \mathbbold{J}^j_{\mathbbold{c}_m^{\scalebox{\hrsup_size}{H}}} &&
		-\mathrm{J}^b_{\mathrm{c}_m^{\scalebox{\hrsup_size}{R}}} && -\mathrm{J}^j_{\mathrm{c}_m^{\scalebox{\hrsup_size}{R}}}
	\end{bmatrix}
	\begin{bmatrix}
		\dot{\mathbbold{v}}_B \\
        \ddot{\mathbbold{s}} \\
        \dot{\mathrm{v}}_B \\
        \ddot{\mathrm{s}}
	\end{bmatrix} = 0
	\label{HRholcon2}
\end{equation}

\end{subequations}

Furthermore, the constraint equations~\eqref{holcon1}~\eqref{holcon2}~\eqref{HRholcon1} and \eqref{HRholcon2} can be represented in a compact form as follows:

\begin{equation}
	\mathbold{P}
	\mathbold{V} +
	\mathbold{Q}
	\dot{\mathbold{V}} = 0
	\label{HRholcon3}
\end{equation}

where,
\begingroup
    \fontsize{8pt}{15pt}\selectfont
    \begin{itemize}
    	\item $\mathbold{P} =
    		   \begin{bmatrix}
    		   		\dot{\mathbbold{J}}^b_{\mathbbold{c}_k} && \dot{\mathbbold{J}}^j_{\mathbbold{c}_k} && 0 && 0\\
    	      		0 && 0 &&
    	      		\dot{\mathrm{J}}^b_{\mathrm{c}_k} && \dot{\mathrm{J}}^j_{\mathrm{c}_k} \\
    	      		\dot{\mathbbold{J}}^b_{\mathbbold{c}_m^{\scalebox{\hrsup_size}{H}}} && \dot{\mathbbold{J}}^j_{\mathbbold{c}_m^{\scalebox{\hrsup_size}{H}}} &&
    	      		-\dot{\mathrm{J}}^b_{\mathrm{c}_m^{\scalebox{\hrsup_size}{R}}} && -\dot{\mathrm{J}}^j_{\mathrm{c}_m^{\scalebox{\hrsup_size}{R}}}
    		   \end{bmatrix} \in \mathbb{R}^{6 \times (\mathbbold{n} + \mathrm{n} + 12)}$ \vspace{0.2 cm}
    	\item $\mathbold{Q} =
    		   \begin{bmatrix}
    				\mathbbold{J}^b_{\mathbbold{c}_k} && \mathbbold{J}^j_{\mathbbold{c}_k} &&
    				0 && 0 \\
    					0 && 0 &&
    				\mathrm{J}^b_{\mathrm{c}_k} && \mathrm{J}^j_{\mathrm{c}_k} \\
    					\mathbbold{J}^b_{\mathbbold{c}_m^{\scalebox{\hrsup_size}{H}}} && \mathbbold{J}^j_{\mathbbold{c}_m^{\scalebox{\hrsup_size}{H}}} &&
    				-\mathrm{J}^b_{\mathrm{c}_m^{\scalebox{\hrsup_size}{R}}} && -\mathrm{J}^j_{\mathrm{c}_m^{\scalebox{\hrsup_size}{R}}}
    		   \end{bmatrix} \in \mathbb{R}^{6 \times (\mathbbold{n} + \mathrm{n} + 12)}$ \vspace{0.2 cm}
    	\item  $\mathbold{V} =
    			\begin{bmatrix}
    				\bbnu && \nu
    			\end{bmatrix}^T \in \mathbb{R}^{\mathbbold{n}+\mathrm{n}+12}$
    \end{itemize}
\endgroup

\subsection{Contact and interaction wrenches}

First, observe that we can combine equation \eqref{NEHuman2} and equation \eqref{NERobot2} to obtain a single equation of motion for the composite system as shown in Eq. \eqref{NE-HR1}

\begin{equation}
    \small
	\begin{bmatrix}
		\mathbbold{M} && 0 \\
		0 && \mathrm{M}
	\end{bmatrix}
	\begin{bmatrix}
		\dot{\bbnu} \\
		\dot{\nu}
	\end{bmatrix} +
	\begin{bmatrix}
		\mathbbold{h} \\
		\mathrm{h}
	\end{bmatrix} =
	\begin{bmatrix}
		\mathbbold{B} && 0 \\
		0 && \mathrm{B}
	\end{bmatrix}
	\begin{bmatrix}
		{\tau}^{\scalebox{\hrsup_size}{H}} \\
		{\tau}^{\scalebox{\hrsup_size}{R}}
	\end{bmatrix} +
	\begin{bmatrix}
		\mathbbold{J}^T && 0 \\
		0 && \mathrm{J}^T
	\end{bmatrix}
	\begin{bmatrix}
		\mathbbold{f}^* \\
		\mathrm{f}^*
	\end{bmatrix}
	\label{NE-HR1}
\end{equation}

\vspace{0.1 cm}
where, $\mathbbold{h} = \mathbbold{C}(\mathbbold{q},\bbnu)\bbnu + \mathbbold{G}(\mathbbold{q})$, $\mathrm{h} = \mathrm{C}(\mathrm{q},\nu)\nu + \mathrm{G}(\mathrm{q})$

According to the Newtonian mechanics, in the case of rigid contacts the perturbations exerted by the robot on the human is equal and opposite to the perturbation exerted by the human on the robot. As a consequence, when the external wrenches are expressed with respect to the inertial frame, the interaction wrenches $\mathrm{f}$ can be written as follows:

\begin{equation}
	\mathrm{f} \ = \ \mathbbold{f}_m \ = -\mathrm{f}_m
	\label{mutual-wrenches}
\end{equation}

As a consequence, the equation \eqref{NE-HR1} can be written in a compact form as follows:
\begin{equation}
	\mathbold{M} \dot{\mathbold{V}} + \mathbold{h} = \mathbold{B} \mathbold{\tau} + \mathbold{J}^T \mathbold{f}^*
	\label{NE-HR2}
\end{equation}

where $\mathbold{f}^* = \begin{bmatrix}
							\mathrm{f} && \mathbbold{f}_e && \mathrm{f}_e
					    \end{bmatrix}^T \in \mathbb{R}^{6 ({\mathbbold{n}_m} + \mathbbold{n}_e + \mathrm{n}_e) }$ and
$\mathbold{J}$ a proper jacobian  matrix. This equation implies that

\begin{equation}
	\dot{\mathbold{V}} = {\mathbold{M}^{-1}}[\mathbold{B}\mathbold{\tau}+\mathbold{J}^T \mathbold{f}^* - \mathbold{h}]
	\label{HR-vel}
\end{equation}

We make use of the equation \eqref{HR-vel} in the constraint equation \eqref{HRholcon3}

\begin{subequations}
	\begin{equation}
		\mathbold{P} \mathbold{V} + \mathbold{Q} {\mathbold{M}^{-1}}[\mathbold{B}\mathbold{\tau}+\mathbold{J}^T \mathbold{f}^* - \mathbold{h}] = 0 \notag
	\end{equation}
	\begin{equation}
		\Rightarrow \mathbold{Q} {\mathbold{M}}^{-1} \mathbold{J}^T \mathbold{f}^* = - [ \mathbold{Q} {\mathbold{M}}^{-1} [\mathbold{B} \mathbold{\tau} - \mathbold{h}] + \mathbold{P} \mathbold{V}] \notag
	\end{equation}
\end{subequations}

\begin{equation}
		\Rightarrow \mathbold{f}^* = -\Gamma^{-1}[ \mathbold{Q} {\mathbold{M}}^{-1} [\mathbold{B} \mathbold{\tau} - \mathbold{h}] + \mathbold{P} \mathbold{V}] \notag
\end{equation}
\noindent where, $\Gamma = \mathbold{Q} {\mathbold{M}}^{-1} \mathbold{J}^T$
\vspace*{0.2 cm} \\
Furthermore,
\begin{equation}
	\mathbold{f}^* = - \Gamma^{-1} \mathbold{Q} {\mathbold{M}}^{-1} \mathbold{B} \mathbold{\tau} +
					   \Gamma^{-1} \mathbold{Q} {\mathbold{M}}^{-1} \mathbold{h} -
					   \Gamma^{-1} \mathbold{P} \mathbold{V}
	\label{external-wrenches}
\end{equation}

\vspace*{0.2 cm}
Through coupled-dynamics, equation \eqref{external-wrenches} shows that the external wrenches are a function of system configuration $\mathbbold{q}$, $\mathrm{q}$, system velocity $\bbnu$, $\nu$, and joint torques $\tau^{\scalebox{\hrsup_size}{H}}$, $\tau^{\scalebox{\hrsup_size}{R}}$. This can be represented as a function $
	\mathbold{f}^* = g(\mathbbold{q},\mathrm{q},\bbnu,\nu, \tau^{\scalebox{\hrsup_size}{H}},\tau^{\scalebox{\hrsup_size}{R}}) \notag
$. This relation can be further decomposed as,

\begin{equation}
	\mathbold{f}^* = \mathbold{G}_1 \tau^{\scalebox{\hrsup_size}{H}} + \mathbold{G}_2 \tau^{\scalebox{\hrsup_size}{R}} + {\mathbold{G}_3}(\mathbbold{q},\mathrm{q},\bbnu,\nu)
	\label{external-wrenches-compact}
\end{equation}

\noindent where,
\begin{itemize}
	\item $\mathbold{G}_1 \in \mathbb{R}^{6 ({\mathbbold{n}_m} + \mathbbold{n}_e + \mathrm{n}_e) \times \mathbbold{n}}$
	\item $\mathbold{G}_2 \in \mathbb{R}^{6 ({\mathbbold{n}_m} + \mathbbold{n}_e + \mathrm{n}_e) \times \mathrm{n}}$
	\item $\mathbold{G}_3 \in \mathbb{R}^{6 ({\mathbbold{n}_m} + \mathbbold{n}_e + \mathrm{n}_e)}$
\end{itemize}

\section{Partner-aware control}
\label{control}

Let $\chi \in \mathbb{R}^p$ be a robot-related quantity of dimension $p$ that is assumed to have a linear map to the robot's velocity, i.e.:
\begin{equation}
	\chi = \mathrm{J}_{\chi}(\mathrm{q}) \ \nu
	\label{chi-jacobian-map}
\end{equation}
where $\mathrm{J}_{\chi}(\mathrm{q}) \in \mathbb{R}^{p \times (\mathrm{n}+6)}$. Let $\chi_d$ denote the desired value of $\chi$, and  $\widetilde{\chi} = \chi - \chi_d$ the error to be minimized. On time differentiating (\ref{chi-jacobian-map}) and by substituting the robot acceleration $\dot{\nu}$ with its expression obtained from the model (\ref{NERobot2}) one can observe that the acceleration $\dot{\chi}$ depends upon the robot torques ${\tau}^{\scalebox{\hrsup_size}{R}}$, namely, $\dot{\chi} = \dot{\chi}({\tau}^{\scalebox{\hrsup_size}{R}})$. Then, a classical approach for the control of the robot quantity $\chi_d$ consists of finding the robot joint torques ${\tau}^{\scalebox{\hrsup_size}{R}}$ such that
\begin{equation}
	\label{closedLoop}
	\dot{\chi} = \dot{\chi}_d - k_d \ \widetilde{\chi} -k_p \ \int_0^t \widetilde{\chi} ds, \quad k_d, k_p > 0
\end{equation}

This is a classical feedback linearization approach \cite{Khalil2004}. In the language of optimization theory, the above feedback linearization control task can be framed in the following optimization problem
\begin{subequations}
\begin{eqnarray}
	{{\tau}^{\scalebox{\hrsup_size}{R}}}^* &=& \arg \min_{{\tau}^{\scalebox{\hrsup_size}{R}}}  |\dot{\chi}({\tau}^{\scalebox{\hrsup_size}{R}}) - \dot{\chi}_d +k_d \widetilde{\chi} +k_p \int_0^t \widetilde{\chi} ds|^2     \\
		   &s.t.& \nonumber \\
		   &&\mathbold{M} \dot{\mathbold{V}} + \mathbold{h} = \mathbold{B} \mathbold{\tau} + \mathbold{J}^T \mathbold{f}^*
     \\
		   &&\mathbold{P} \mathbold{V} + \mathbold{Q} \dot{\mathbold{V}} = 0
\end{eqnarray}
\end{subequations}

The feedback linearization approach is fundamentally agnostic to any interaction from an external agent. This is evident from Eq.~\eqref{closedLoop} since no human quantity appears on the right-hand side of this equation. This motivates us to propose partner-aware robot control techniques that \emph{exploit} help provided by an external agent during the physical interaction.

\subsection{Partner-aware robot control leveraging external help}

Certainly, instead of completely canceling out any external interaction by the feedback control action, it is gainful and desirable to \emph{exploit} it to accomplish the robot's task. This poses, however, the question of characterizing and quantifying human help with respect to the robot task. In our previous work \cite{8093992} we relied on using the interaction wrench \textit{estimates} from the robot as human intent information. However, in a coupled system, wrench information introduces an algebraic loop in the control design as the wrench estimates from the robot are computed using the robot joint torques \cite{Frontiers2015}. Instead, a sound approach is to leverage the joint torques of the human as they are largely self-generated and self-regulated. Additionally, considering the joint torques opens new possibilities for our future work to investigate and optimize human ergonomics. Refer to the APPENDIX section \ref{proof:lemma-1} for details on how the human joint torques are exploited in our approach.

The following lemma proposes control laws that exploit the human contribution towards the achievement of the robot control objective, thus actively taking into account the physical human-robot interaction. The associated analysis is based on considering the \emph{energy} of the robot control task.

\begin{lemma}
	Assume that the control objective is to asymptotically stabilize the following point
	\begin{equation}
		\left(\widetilde{\chi},\int\widetilde{\chi} ds\right) = (0,0)
		\label{eq-point}
	\end{equation}
	Apply the following robot torques to the robot system~\eqref{NERobot2}
	\begin{equation}
		\tau^{\scalebox{\hrsup_size}{R}} = - \mathbold{\Delta}^{\dagger} \ [ \ \Lambda \ + \
										 K_D \ \widetilde{\chi} \ + \ max(0,\alpha) \ \widetilde{\chi}^{\parallel} \ ] + N_{\mathbold{\Delta}} \tau^{\scalebox{\hrsup_size}{R}}_0
		\label{control-law}
	\end{equation}
	with
		\begin{itemize}\setlength\itemsep{0.8em}
		    \item $ \mathbold{\Delta} = K_d \ \mathrm{J}_{\chi} \mathrm{M}^{-1} [\mathrm{B} + \mathrm{J}^T \bar{\mathbold{G}}_2 ] \in \mathbb{R}^{p \times \mathrm{n}}$
		    \item $ N_{\mathbold{\Delta}}$ the null-space projector of the matrix $\mathbold{\Delta}$;
		    \item $\tau^{\scalebox{\hrsup_size}{R}}_0$ a free $n-$dimensional vector;
		    \item $ \mathbold{\Lambda} = K_d \ [ \ [ \ \mathrm{J}_{\chi} \mathrm{M}^{-1} \mathrm{J}^T \ ] {\bar{\mathbold{G}}_3}(\mathbbold{q},\mathrm{q},\bbnu,\nu) -     					  		    \mathrm{J}_{\chi} \mathrm{M}^{-1} \mathrm{h} \ + \
						  			\dot{\mathrm{J}}_{\chi} \nu + K_p \ \int_{0}^{t}(\chi - \chi_d) ds - \dot{\chi}_d \ ] \in \mathbb{R}^{p}$
		    \item $K_D \in \mathbb{R}^{p \times p}$ is a symmetric, positive-definite matrix
		    \item $\alpha \in \mathbb{R}$ is a component proportional to the human joint torques $\tau^{\scalebox{\hrsup_size}{H}}$ projected along $\widetilde{\chi}^{\parallel}$ i.e., the direction parallel to $\widetilde{\chi}$
	\end{itemize}
	Assume that the matrix $\mathbold{\Delta}$ is full rank matrix $\forall \ t \in \mathbb{R}^{+}$. Then
	\begin{itemize}\setlength\itemsep{1em}
		\item The trajectories $(\widetilde{\chi},\int\widetilde{\chi} ds)$ are globally bounded
		\item The equilibrium point \eqref{eq-point} is stable
	\end{itemize}
\label{lemma-1}
\end{lemma}

A sketch of the proof of \textit{Lemma \ref{lemma-1}} is outlined in the appendix. Observe that the control law \eqref{control-law} includes a degree of redundancy under the assumption that the matrix $\Delta$ is fat, i.e. the dimension of the robot control task is lower than the robot actuation. As a consequence, the free vector $\tau^{\scalebox{\hrsup_size}{R}}_0$ can be used for other control purposes like robot postural task.

\section{EXPERIMENTS}
\label{experiments}

In the case of complex humanoids robots, state-of-the-art whole-body controllers often consider controlling the robot momentum \cite{7803266} \cite{nava2016stability} and accordingly the Eq. \eqref{chi-jacobian-map} becomes,

\begin{equation}
	\mathrm{L} = \mathrm{J}_{cmm}(\mathrm{q}) \ \nu
	\label{Robot-momentum}
\end{equation}

\noindent where, $\mathrm{J}_{cmm}$ is the centroidal momentum matrix.

\begin{figure*}[h!]
  \centering
  \includegraphics[scale=0.18]{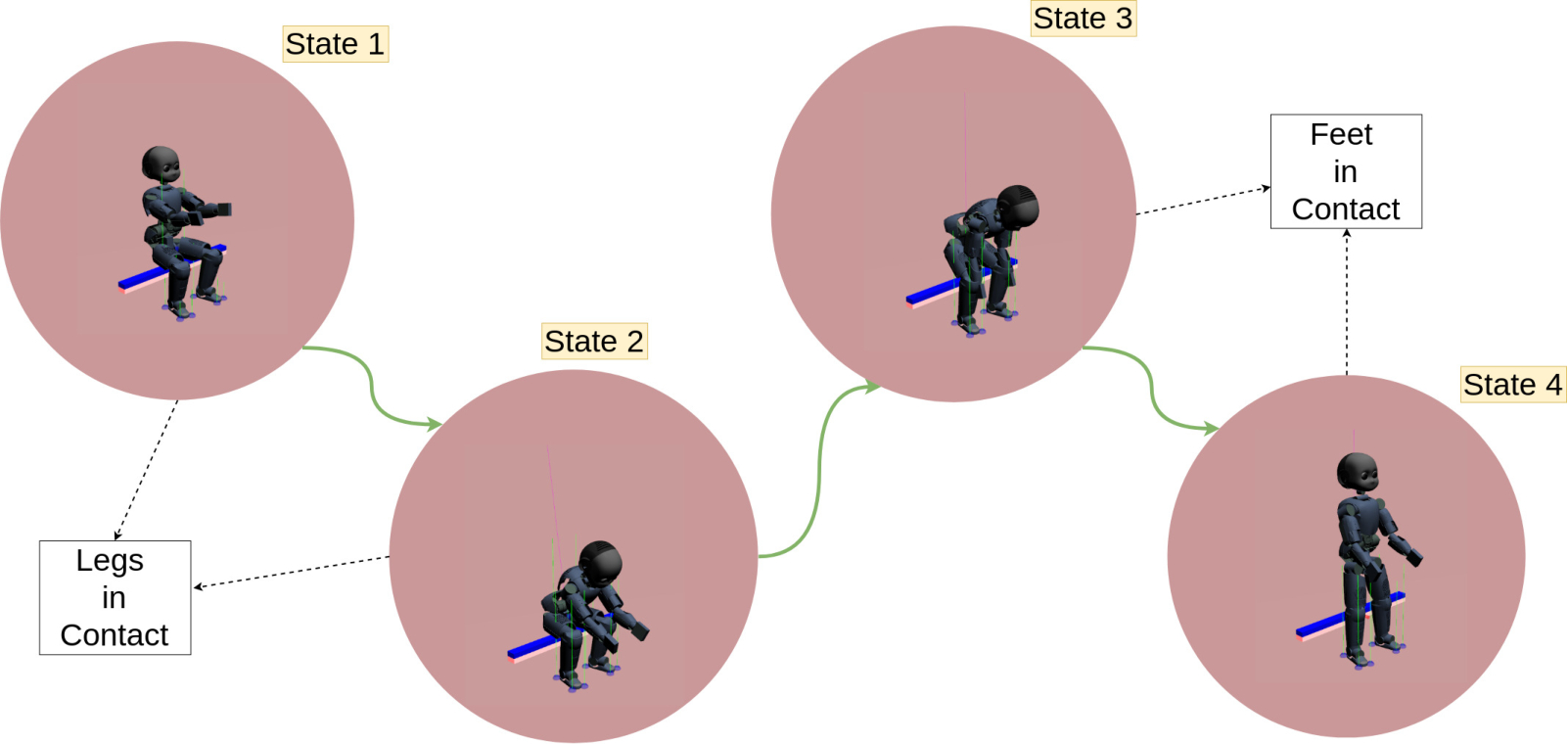}
  \caption{State machine of the controller}
  \label{standup-states}
\end{figure*}

\vspace{0.2cm}
The primary task of the robot we considered is to perform a stand-up motion by moving its center of mass (CoM) through the sit-to-stand transition with \textit{momentum control} as the primary control objective. The robotic platform used in our experiments is the iCub humanoid robot \cite{metta2010icub} \cite{Nataleeaaq1026}. The Simulink controller is designed with four states for the robot as highlighted in  Fig. \ref{standup-states}. During the state 1, the robot balances on a chair and enters to state 2 when an interaction wrench of a set threshold is detected at the hands indicating the start of pull-up assistance from an external agent. During state 2, the robot moves its center of mass forward and enters state 3 when the external wrench experienced at the feet of the robot is above a set threshold. During state 3, the robot moves its center of mass both in forward and upward directions and enters state 4 when the external wrench experienced at the feet of the robot are above another set threshold. Finally, during state 4 the robot moves its center of mass further upward to stand fully erect on both the feet. Accordingly, during the states 1 and 2 the contacts with the environment (chair), accounted in the controller, are at the \emph{upper legs} of the robot. Similarly, for the states 3 and 4 the \emph{feet} contacts of the robot with the environment (ground) are accounted in the controller.

\begin{figure}[h!]
  \centering
  \includegraphics[scale=0.1]{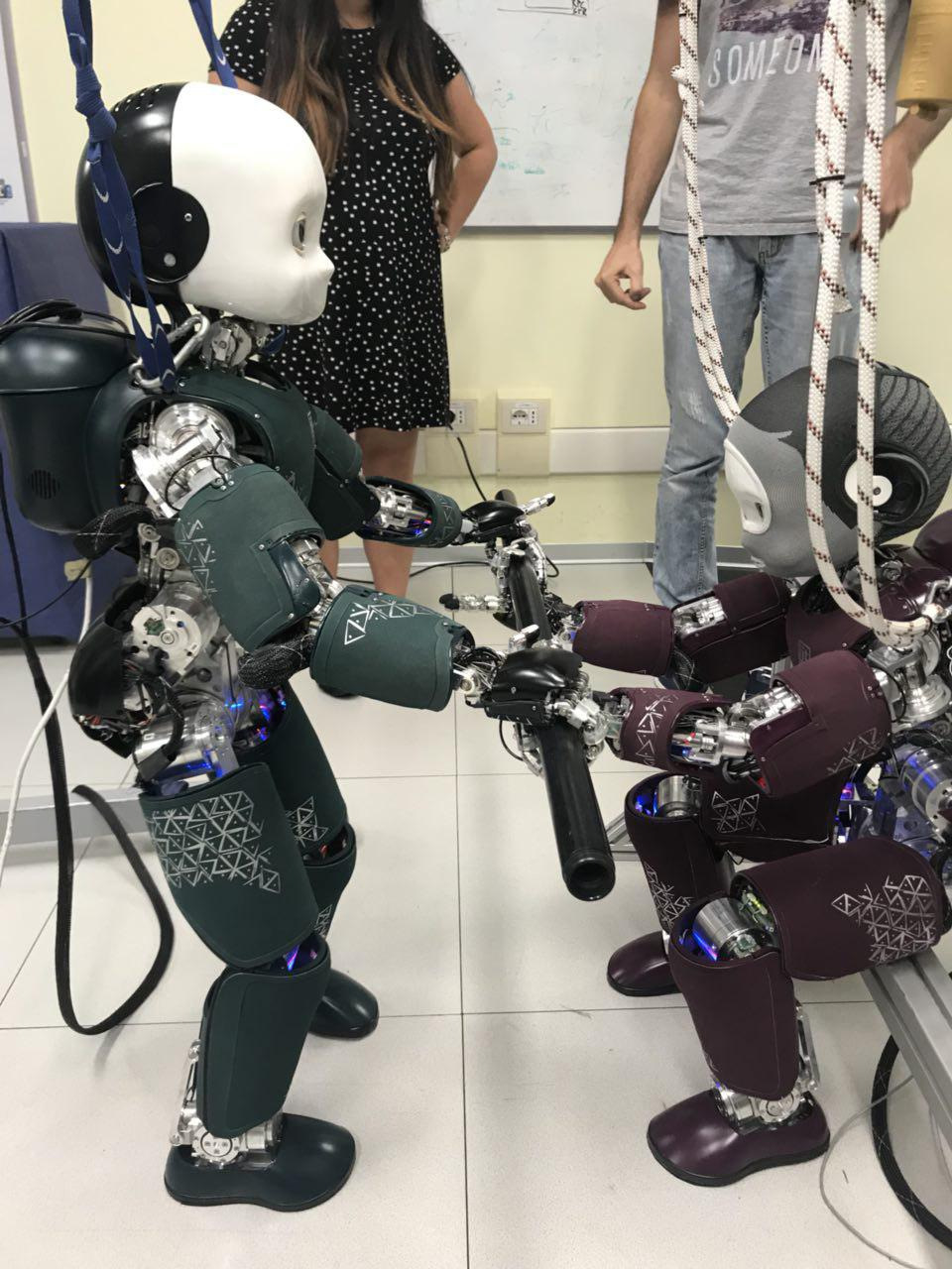}
  \caption{Experimental scenario with two icub robots involved in physical interaction (picture is only representative)}
  \label{two-icubs-pipe}
\end{figure}

Considering the human model as a multi-body mechanical system of rigid links allows us to use another humanoid robot in place of a human for the experiment without the loss of integrity of the experiment to validate our framework. So, we designed a preliminary experimental scenario with two iCub humanoid robots as shown in Fig. \ref{two-icubs-pipe}. 

\begin{figure}[h!]
    \hspace{-1cm}
    \includegraphics[scale=0.2]{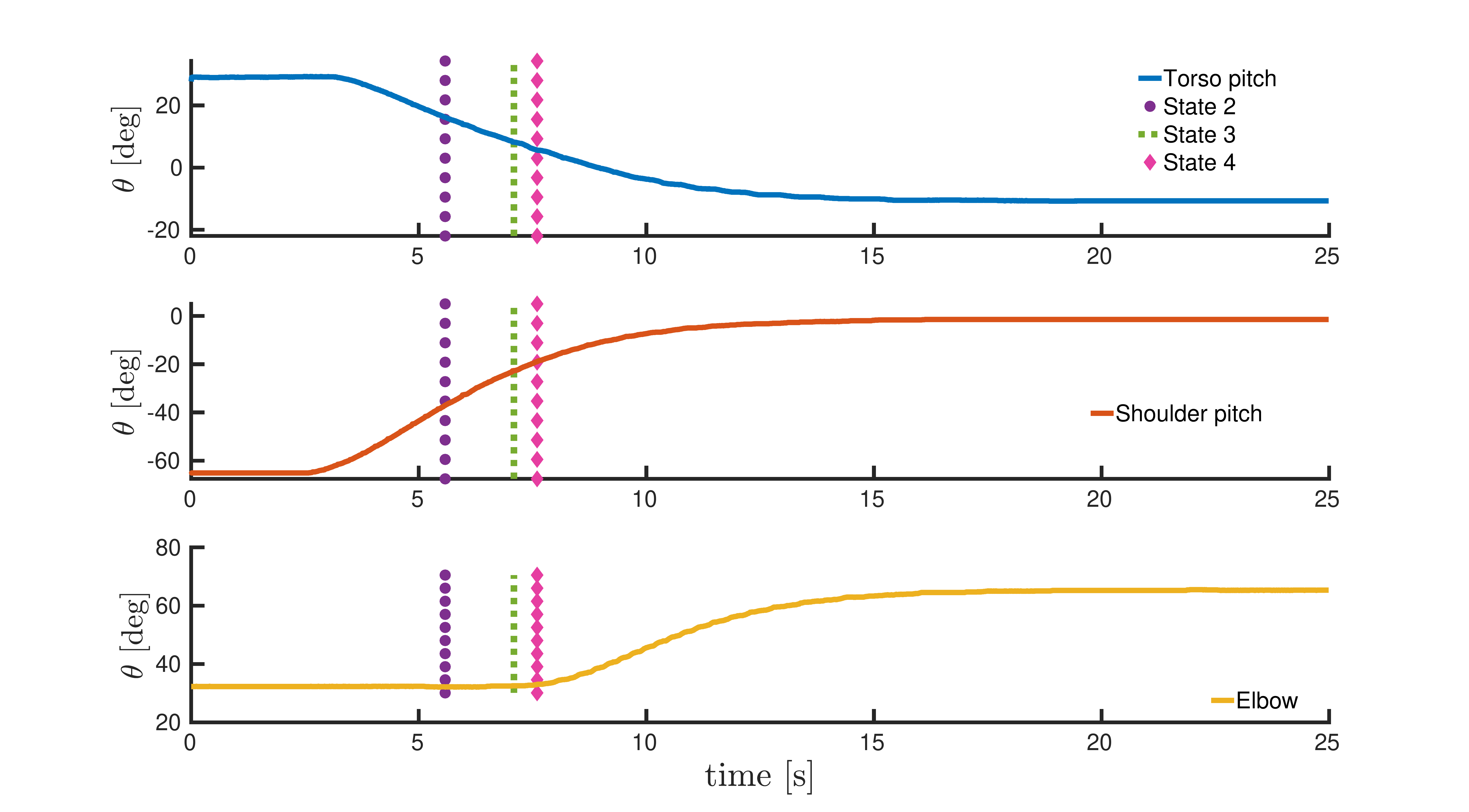}
    \caption{Interacting agent joint trajectories}
    \label{humanQ}
\end{figure}

\begin{figure}[h!]
    \hspace{-1cm}
    \includegraphics[scale=0.2]{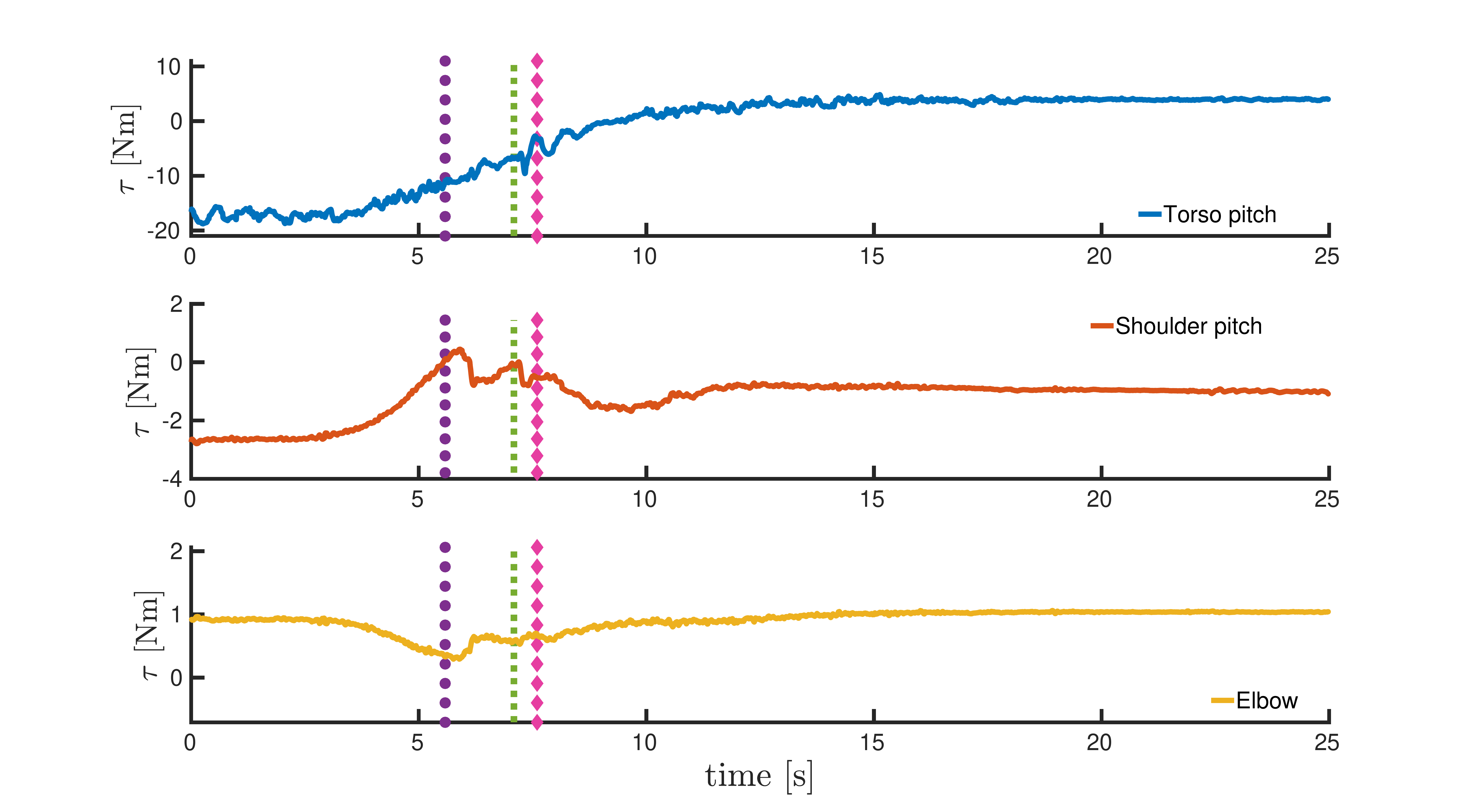}
    \caption{Interacting agent joint torques}
    \label{humanTau}
\end{figure}

The \emph{purple} iCub robot is run in torque control mode and received torque inputs from the controller for the stand-up task. The \emph{green} iCub robot is in position control mode. A predetermined trajectory generated using a minimum-jerk trajectory generator \cite{5650851} is given as a reference to the torso pitch, shoulder pitch and elbow joints of the \emph{green} iCub robot. The resulting motion mimics the pull-up assistance to the \emph{purple} iCub robot for performing the stand-up task. Hence, the \emph{green} iCub robot is considered as an external interacting agent whose joint motion is indicated in Fig. \ref{humanQ} and the associated joint torques are shown in Fig. \ref{humanTau}.

The hands of the iCub robot are quite fragile and are not designed to make \emph{sustained} power grasps. This posed quite a challenge during our experiments. Mechanical coupling of the hands to the bar using just the fingers is hard to maintain rigidly during the entire duration of the experiment because of the hardware limitations of the fingers motors. So, we took additional precautions of using tape to ensure strong coupling throughout the experiment.

\section{RESULTS}
\label{results}

A predetermined trajectory generated using a minimum-jerk trajectory generator \cite{5650851} is given as a reference to the center of mass of the \emph{purple} iCub robot to perform the sit-to-stand transition. At the start of the Simulink controller the \emph{purple} iCub robot is seated on the metallic structure that serves as a chair. Once the controller is started, it receives joint quantities as inputs from both the robots and actively generates joint torque inputs for the \emph{purple} iCub robot to maintain its momentum and track the center of mass.

\begin{figure}[h!]
    \hspace{-1.2cm}
	\includegraphics[scale=0.2]{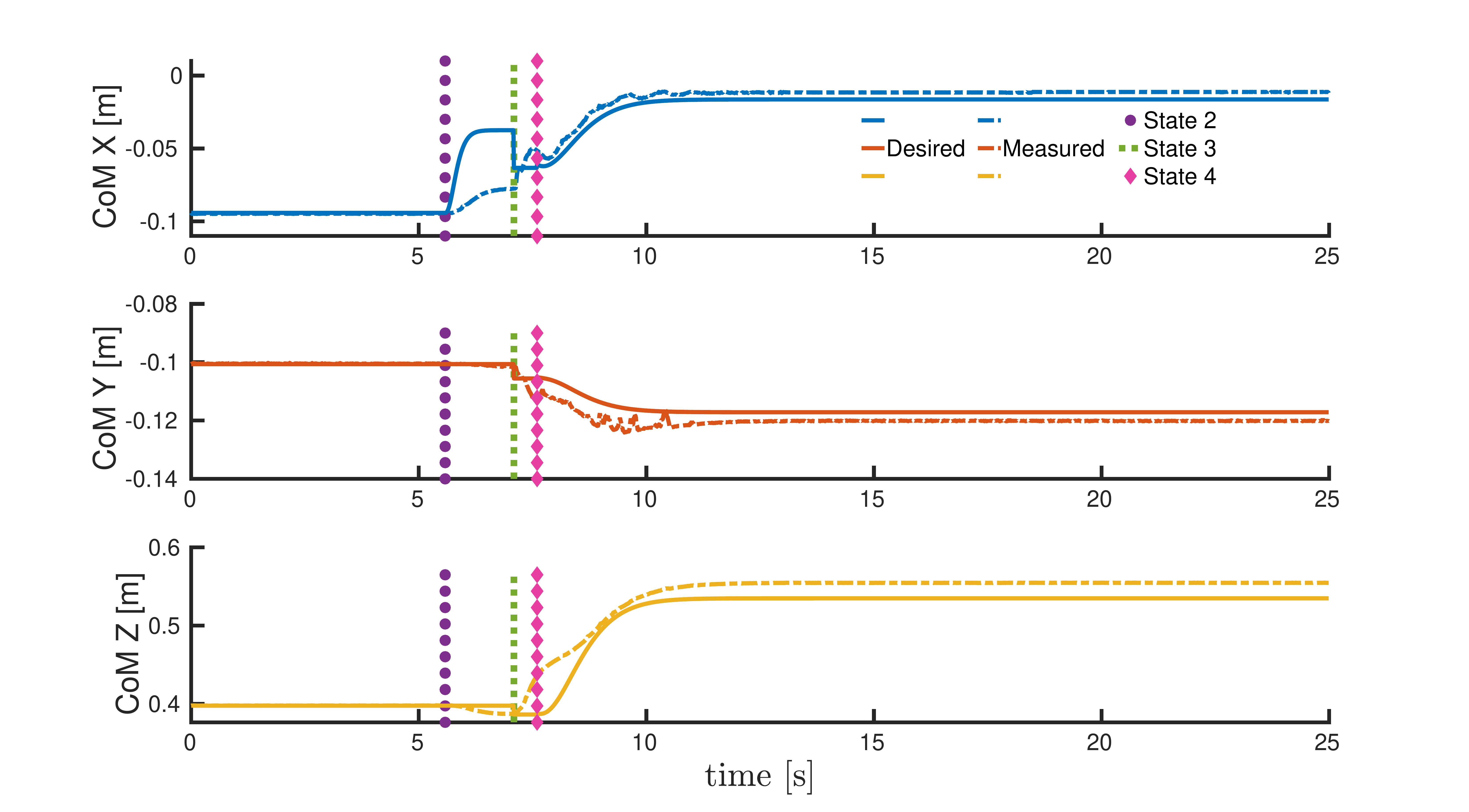}
	\caption{Time evolution of the desired and measured CoM trajectory while performing stand-up motion on application of the control law \eqref{control-law}}
	\label{comErr}
\end{figure}

The time evolution of the center of mass tracking is shown in Fig. \ref{comErr}. The vertical lines indicate the time instance at which a new state begins. Between states 2 and 3, the \emph{purple} iCub robot is seated on the chair with its upper leg as contact constraints. This seriously limits the robot motion along the $x$-direction and hence the tracking error of the center of mass along the $x$-direction is not negligible. Similarly, between states 3 and 4, the robot has to stand-up relative quickly and the contact constraints change from the upper legs to the feet. This contact switching, along with unmodeled phenomena such as joint friction limits the robot motion along the $z$-direction and hence the tracking error of the center of mass along the $z$-direction is not negligible. Apart from these two instances, the overall center of mass tracking is good.

\begin{figure}[h!]
	\hspace{-1.2cm}
	\includegraphics[scale=0.2]{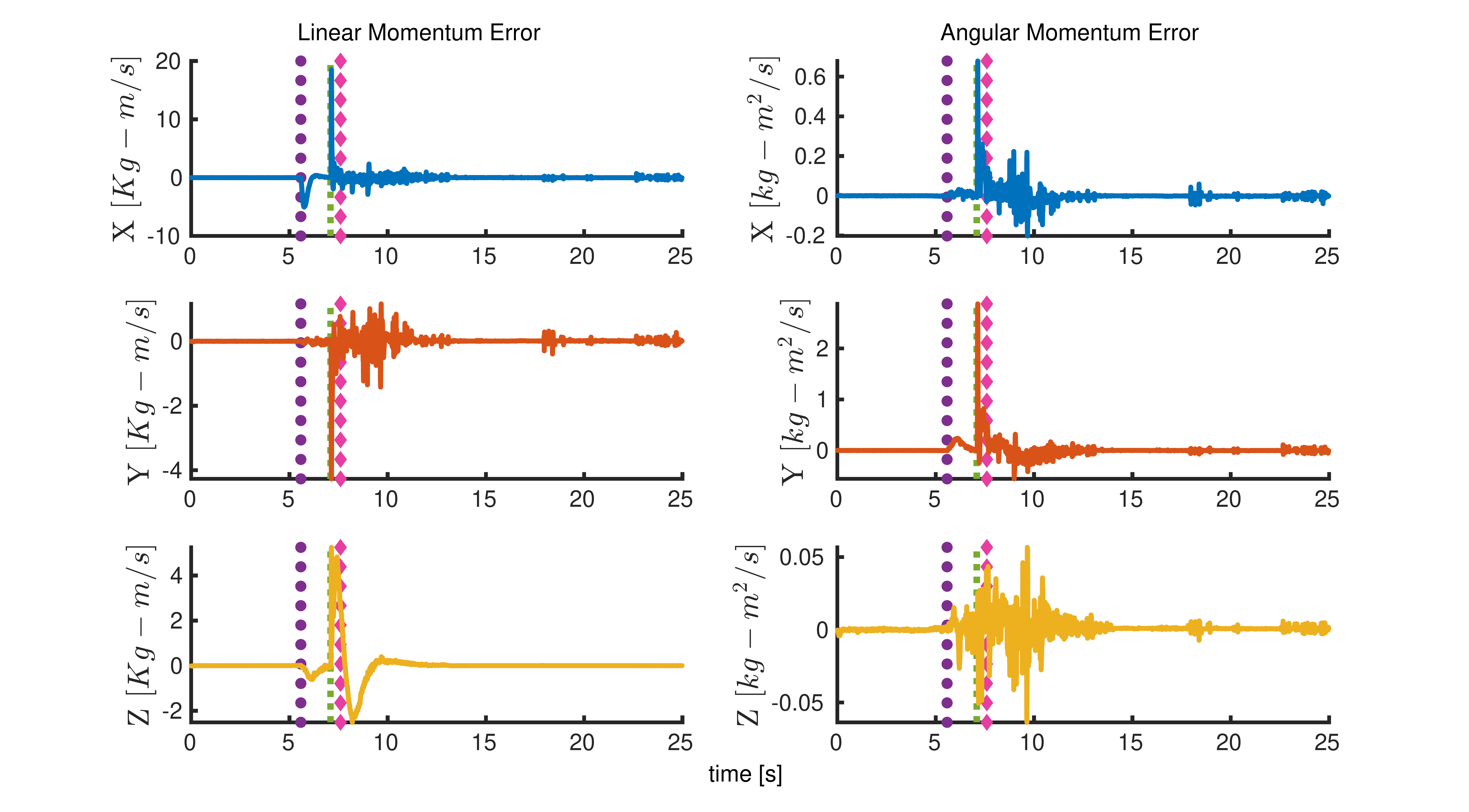}
	\caption{Time evolution of the linear and angular momentum while performing stand-up motion on application of the control law \eqref{control-law}}
	\label{momentumErr}
\end{figure}

The primary control objective of momentum control is also realized well as highlighted by the time evolution of the linear and angular momentum of the robot as shown in Fig. \ref{momentumErr}. Between states 3 and 4, both the linear and angular momentum error increased momentarily. Understandably this results from the impact at the contact switching that occurs at the beginning of state 3. However, the overall robot momentum is maintained close to zero and eventually, the momentum error converges to zero when the robot becomes stable after standing fully erect.

\begin{figure}[h!]
	\hspace{-1cm}
	\includegraphics[scale=0.195]{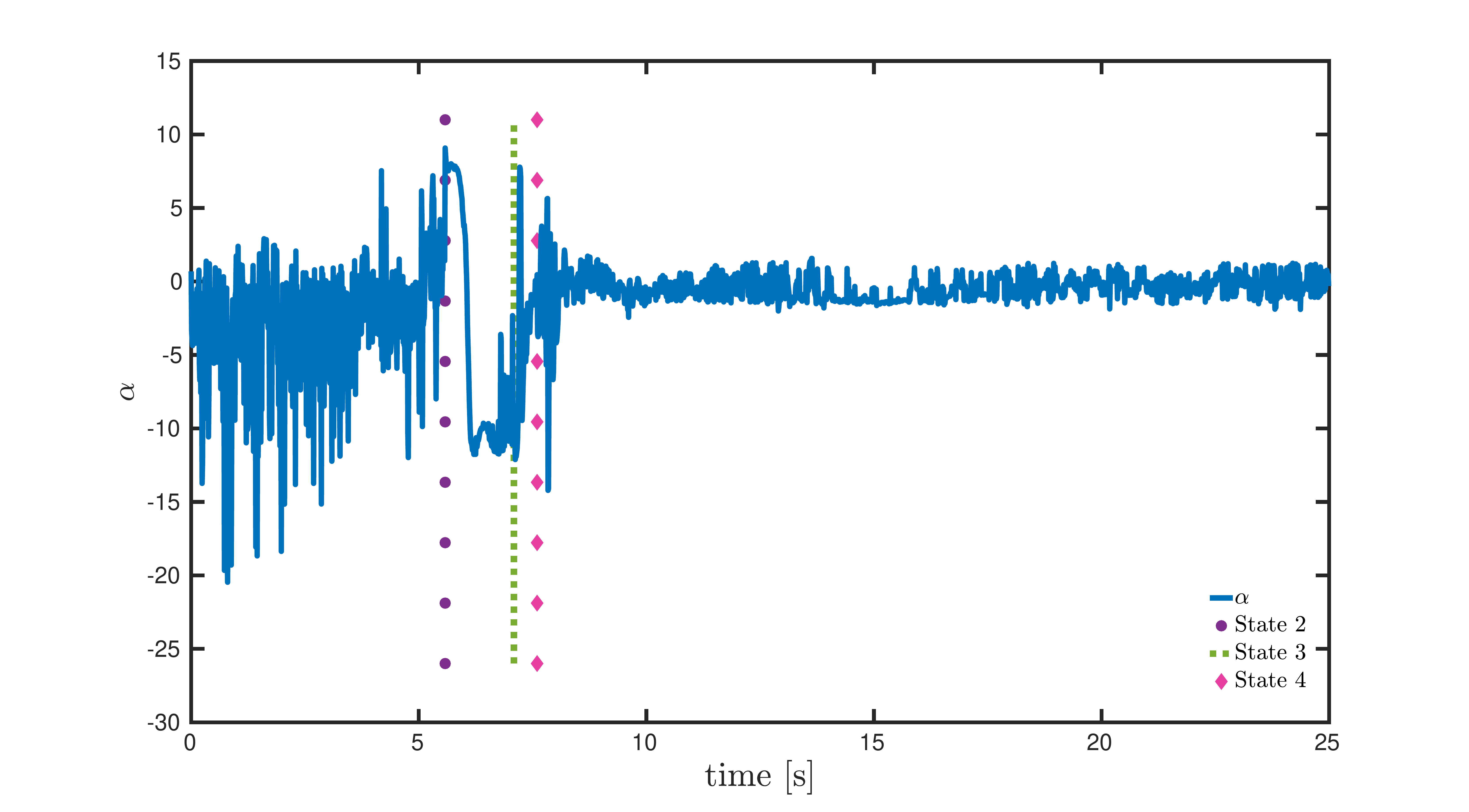}
	\caption{Time evolution of $\alpha$ under the influence of physical interaction with external agent}
	\label{alpha}
\end{figure}

\begin{figure}[h!]
	\hspace{-1.2cm}
	\includegraphics[scale=0.27]{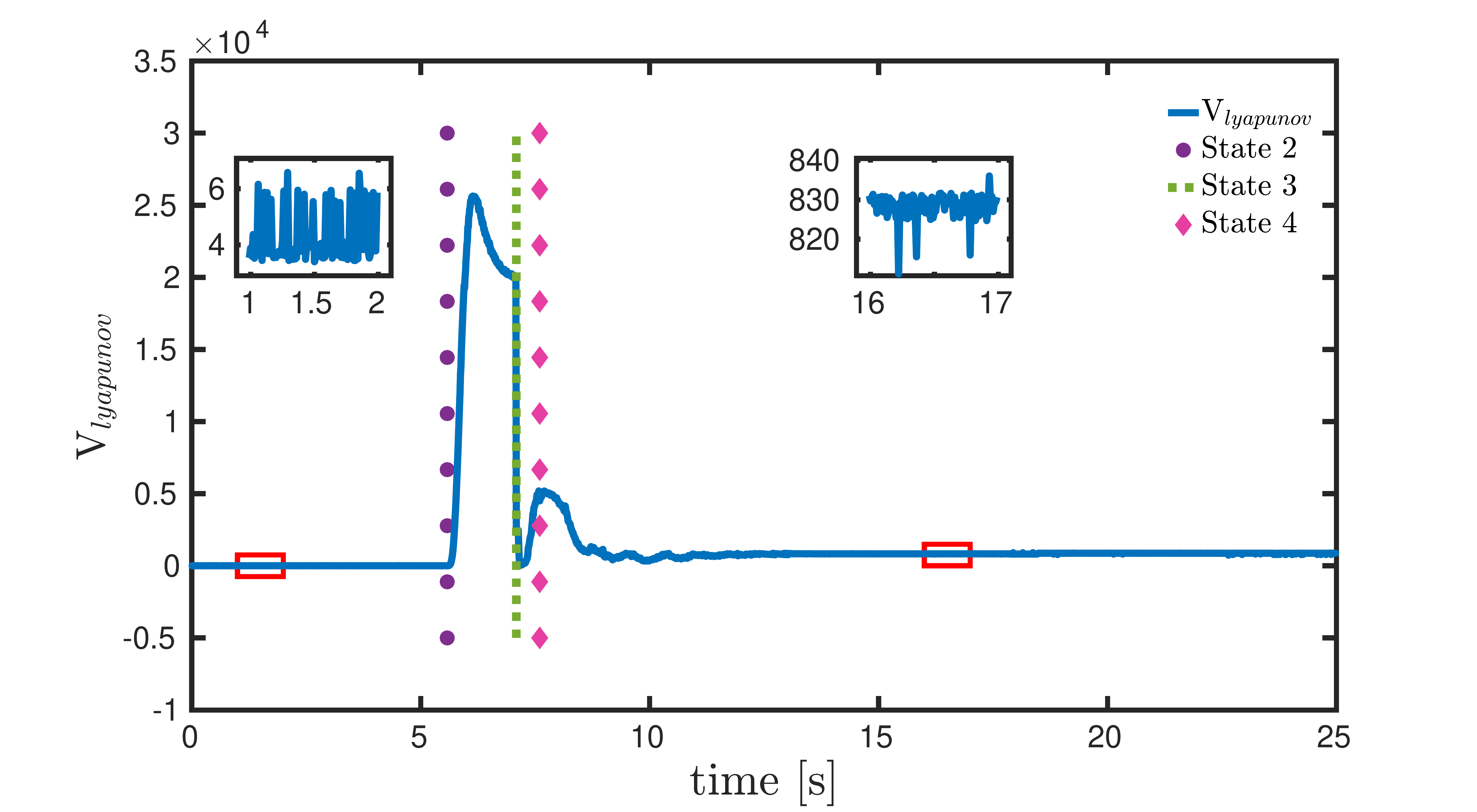}
	\caption{Time evolution of lyapunov function on application of the control law \eqref{control-law}}
	\label{vLyap}
\end{figure}

The time evolution of $\alpha$ i.e. the component of the interaction agent joint torques projected in the direction parallel to the task is shown in Fig. \ref{alpha}. The instantaneous values of $\alpha$ change over the course of the experiment according to the joint torque values of the \emph{green} iCub robot. Between the states 2 and 3, the negative values of $\alpha$ contribute towards making the Lyapunov function decrease faster as indicated in Fig. \ref{vLyap}. This highlights the fact that the physical interaction with an external agent is exploited (in terms of the joint torques) by the \emph{purple} iCub robot to perform the stand-up task. 

The time evolution of the Lyapunov function $\mathrm{V}$ from equation (\ref{LyapunovFunction}) is shown in Fig. \ref{vLyap}. After the controller is started, during the state 1 the system has small energy while the robot is seated on the chair. This is highlighted in the inset plot shown for the duration between 1-2 seconds. Starting from state 2 as the robot starts moving, the total energy of the systems starts to increase as shown between states 2 and 3. As the robot enters state 3 the energy quickly drops during the contact switching from upper legs to feet. This is a direct reflection of exploiting the physical interaction with the \emph{green} iCub robot. Between state 3 and 4 while the robot is moving to a fully erect stance the energy rises slightly and eventually settles to a stable value. The inset plot during the duration between 16-17 seconds highlights the system energy when the robot is in a stable fully erect position. Clearly, the energy needed for a full stand-up posture is higher than that of the seated position during the state 1.

\section{CONCLUSIONS AND FUTURE WORK}
\label{conclusions}

In this paper, we highlighted how the classical approach of feedback linearization fails short to account for the physical interactions from an external agent. Then we presented a generalized framework for human-robot interaction based on coupled-dynamics approach and attempted at mathematically formalizing help from an external agent like a human during physical interaction with the robot. Additionally, we presented a sound approach to consider human intent in the case of a coupled system in order to gainfully exploit the interaction to achieve the task. We built a partner-aware controller and validated our approach by realizing the complicated task of standing up involving physical interactions between two complex humanoid robots. The preliminary case study with two humanoid robots clearly demonstrates the general applicability of our framework and proves that the resulting controllers are both reactive and robust to external physical interaction according to the nature of the interaction with respect to the task at hand.

In the near future, we will conduct experiments involving a human subject in a sensorized suit providing a real-time estimation of the human dynamics \cite{latella2018towards} and present the results with full proof of stability and convergence of control laws. Furthermore, we are investigating the possibilities to improve human ergonomy using the assistance from the robotic agent involved in the physical interaction.

\section*{APPENDIX: sketch of proof of Lemma~\ref{lemma-1}}
\label{proof:lemma-1}

\noindent
\textbf{Proof:} The stability of $\widetilde{\chi}$ can be analyzed by considering the following Lyapunov function:

\begin{equation}
	\label{LyapunovFunction}
	\mathrm{V} = \frac{K_d}{2} \norm{\chi - \chi_d}^{2} + \frac{K_p}{2} \norm{\int_{0}^{t}(\chi - \chi_d) ds}^{2}
\end{equation}

where $K_d, K_p \in \mathbb{R}^{p \times p}$ are two symmetric, positive-definite matrices. Now, on differentiating \eqref{LyapunovFunction} and using the robot dynamics \eqref{NERobot2} along with the force decomposition \eqref{external-wrenches-compact} obtained through coupled-dynamics, we get:

\begin{equation}
	\dot{\mathrm{V}} =
	 - \
    \widetilde{\chi}^T \ K_D \ \widetilde{\chi} \ + \widetilde{\chi}^T \ [ \ \alpha - \
    max(0,\alpha)\ ] \ \widetilde{\chi}^{\parallel}
    \label{Vdot-control}
\end{equation}

where, 
\begin{subequations}

    \begin{equation}
        \dot{\mathrm{V}} =
	 - \
    \widetilde{\chi}^T \ K_D \ \widetilde{\chi} \quad \quad \forall \ \alpha > 0 \notag
    \end{equation}
    
    \begin{equation}
        \dot{\mathrm{V}} =
	 - \
    \widetilde{\chi}^T \ K_D \ \widetilde{\chi} \ +  \widetilde{\chi}^T \ \alpha \ \widetilde{\chi}^{\parallel} \quad \quad \forall \ \alpha \le 0 \notag
    \end{equation}
    
\end{subequations}

The fact that the human joint torques help the robot to perform a control action is encompassed in the right-hand side of the above equation. The component of human joint torques projected in the direction parallel to the task i.e. $\alpha$ makes the Lyapunov function decrease faster. Thus the control law (\ref{control-law}) ensures that $\dot{\mathrm{V}} \leq 0$ which proves that the trajectories are globally bounded. From Lyapunov theory, as $\dot{\mathrm{V}} \leq 0 $ in the neighborhood of the point $(0,0)$ the equilibrium point (\ref{eq-point}) is stable. The complete proof of \textit{Lemma \ref{lemma-1}} is beyond the scope of this paper due to the space limitations and it will be presented in full in our forthcoming journal publication.

\section{ACKNOWLEDGEMENTS}
This work is supported by \href{http://itn-pace.eu/}{PACE} project, Marie Skłodowska-Curie grant agreement No. 642961 and \href{https://andy-project.eu/}{An.Dy} project which has received funding from the European Union\textquotesingle s Horizon 2020 Research and Innovation Programme under grant agreement No. 731540. The authors would like to thank Yue Hu, Stefano Dafarra, Giulio Romualdi and, Aiko Dinale for their support in conducting experiments.

%
%

\bibliographystyle{plain}
\bibliography{main}

\end{document}